\documentclass[runningheads]{llncs}
\usepackage{graphicx}
\usepackage{epsfig}
\usepackage{amsmath,amssymb} 
\usepackage{color}
\usepackage{multirow}
\usepackage{adjustbox}
\usepackage{hyperref}
\usepackage{siunitx}
\usepackage{subfig}
\usepackage[width=122mm,left=12mm,paperwidth=146mm,height=193mm,top=12mm,paperheight=217mm]{geometry}
\begin{document}

\newcommand{\todo}[1]{ \textcolor{red}{\bf #1}}

\newcommand{\fsEq}{0.50}
\newcommand{\fsSm}{0.8}
\newcommand{\fsBg}{0.8}
\newcommand{\fsSmUCF}{0.15}
\newcommand{\fsBgUCF}{0.30}
\newcommand{\fsSmUCFst}{0.25}
\newcommand{\fsBgUCFst}{0.50}

\makeatletter
\def\blfootnote{\gdef\@thefnmark{}\@footnotetext}
\makeatother

\pagestyle{headings}
\mainmatter
\def\ECCV16SubNumber{1840}  

\title{Predicting the Future with Transformational States} 


\titlerunning{Predicting the Future with Transformational States}
\authorrunning{Jaegle, Rybkin, Derpanis, and Daniilidis}
\author{Andrew Jaegle\textsuperscript{1}, Oleh Rybkin\textsuperscript{1}, Konstantinos G.\ Derpanis\textsuperscript{2}, \\and Kostas Daniilidis\textsuperscript{1}}
\institute{\textsuperscript{1}University of Pennsylvania, \textsuperscript{2}Ryerson University \\
ajaegle@upenn.edu, oleh@cis.upenn.edu, \\
kosta@scs.ryerson.ca, kostas@cis.upenn.edu}

\maketitle

\begin{abstract}
An intelligent observer looks at the world and sees not only what is, but what is moving and what can be moved. In other words, the observer sees how the present state of the world can transform in the future. We propose a model that predicts future images by learning to represent the present state and its transformation given only a sequence of images. To do so, we introduce an architecture with a latent state composed of two components designed to capture (i) the present image state and (ii) the transformation between present and future states, respectively. We couple this latent state with a recurrent neural network (RNN) core that predicts future frames by transforming past states into future states by applying the accumulated state transformation with a learned operator. We describe how this model can be integrated into an encoder-decoder convolutional neural network (CNN) architecture that uses weighted residual connections to integrate representations of the past with representations of the future.  Qualitatively, our approach generates image sequences that are stable and capture realistic motion over multiple predicted frames, without requiring adversarial training. Quantitatively, our method achieves prediction results comparable to state-of-the-art results on standard image prediction benchmarks (Moving MNIST, KTH, and UCF101).

\end{abstract}

\blfootnote{\url{https://daniilidis-group.github.io/transformational_states}}

\section{Introduction}
Humans and other animals are able to reason about the future state of the world given visual observations of the present. Even as infants, humans can use images to make informed predictions of how objects and agents will move and act in the future \cite{Spelke_review}. A large body of evidence from the neural and cognitive sciences suggests that humans build predictive models of the world and use the resulting predictions to guide action and to learn better ways of engaging with the world \cite{bubic_2010}. The world is filled with image sequences, and it is clear that intelligent agents might use the rich dynamics of visual stimuli to guide learning. But how agents should learn to predict effectively remains an important computational problem.

\begin{figure}[t]
\begin{center}
	\epsfig{file=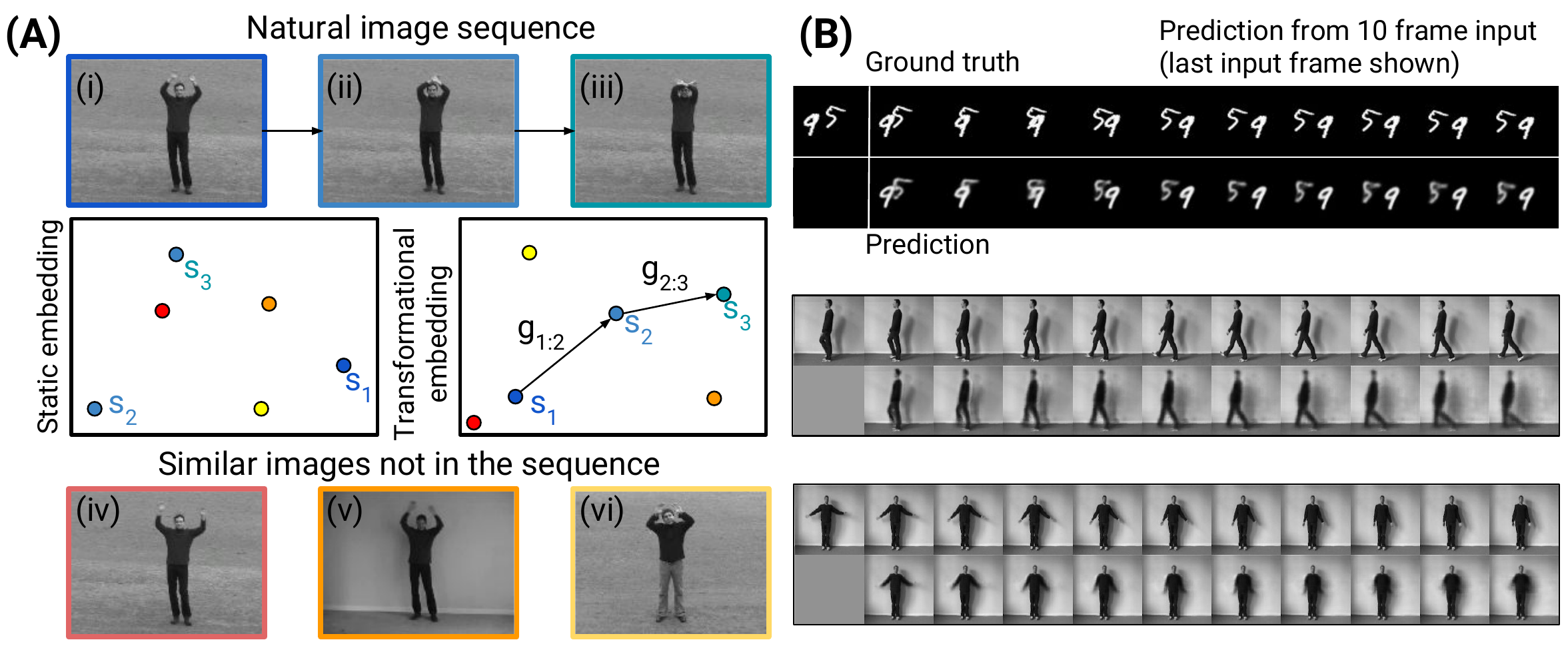,width=5in}
\end{center}
\caption{(A) This work is motivated by the observation that image transformations may be more easily modeled by a network that learns to transform latent states rather than transform or associate pixel intensities. By learning to model states, $s$, along with transformations between states, $g$, the RNN is encouraged to model sequences not by memorizing arbitrary transitions between certain images (static embedding) but by reshaping the embedding so that natural state transformations are predictable (transformational embedding). Figure best viewed in color. (B) Sample predictions of our model on sequences from the Moving MNIST and KTH datasets. We show only the last of 10 input images for visualization purposes. Our model produces good image predictions using only pixel-wise reconstruction losses.}
	\label{fig:overview}
\end{figure}

The focus of this paper is the prediction of future images
given a sequence of past images. Image prediction offers a general approach to tackling the challenge of state prediction in vision because it is not tied to a specific task or representation. By predicting images instead of task-dependent representations such as labels or segmentations, the agent gains more flexibility in how it uses information about the future. From this perspective, image prediction offers a unique opportunity for unsupervised visual representation learning, as image-level predictions can be used as a learning signal even in the absence of well-defined tasks or task-conditional reward signals. 


Motion prediction is at the heart of future prediction in temporally contiguous video. Over short periods of time, scenes in the real world contain a slowly changing context and set of objects. Future frames can largely be predicted by modeling the motion of objects and the scene: that is, by transforming the current state into future states. Our work is motivated by the observation that learning states that can be transformed to produce future states may lead to representations that are easier to predict, as illustrated in Figure \ref{fig:overview}(A). Other methods that use motion for prediction typically rely on the assumption that image transformations can be modeled with local, piecewise translational motion. However, such methods struggle with scenes containing flexible objects like human bodies (e.g. the one shown in Figure \ref{fig:overview}(A)), due to the self-occlusion and non-rigid deformations that such objects introduce. Here, we propose a model that makes predictions by transforming latent representations, and which can reason about transformations that are more complex than simple pixel translations.

To predict realistic images, a model of sequence transformations must also capture the appearance and texture of the scene as it transforms. This includes the content of image regions that appear or become dis-occluded over time. A model must capture the appearance of the foreground and background to paint in details of image regions that are revealed as the objects and scene move. A useful future image prediction model needs to model both this static state and its transformations to ensure that individual frames are realistic and that objects and the scene move realistically. Here, we show how to integrate weighted residual connections into our network to produce good models of background texture and other static scene content.

We propose to predict a latent state representation that encodes both the current state of the scene and its transformation and that can be decoded to produce future images. Our method learns representations of states and transformations that are stable and sufficiently rich to produce multiple future frames without re-encoding estimated frames or repeatedly copying pixels from the input sequence. Our architecture learns to capture naturalistic motion in a variety of settings (synthetic and real) with minimal modifications. Our model achieves quantitative results competitive with the state of the art without assuming a static background (or stabilized preprocessing), without being constrained to directly copying or translating pixels from input frames, and without adversarial learning.

Our technical contributions are as follows: 
\begin{itemize}
\item a novel RNN core formulation with a partitioned representation of latent states and transformations;
\item a weighted, temporal residual connection that enables stably reconciling features across multiple time steps without re-encoding images;
\item an encoder-decoder architecture that can be stably trained for good end-to-end image prediction without an adversarial loss.
\end{itemize}

\section{Related work}
There is a growing interesting in predicting future imagery conditioned on past observations.  This body of work leverages large, unlabeled video datasets to learn to make predictions.  Prior work has explored a variety of aspects of the problem.  Here, we present an overview of prior work organized by the level of abstraction of the target output, the generative process, the structure of the latent representation, and the loss function guiding learning. To further aid interpretation of the present work, we compare the specific design choices of a range of recently proposed models in Supplementary Table \ref{arch-comparisons}.

\textbf{Prediction targets} At the prediction output level, a variety of representational levels have been targeted.  At the highest level, several works have considered predicting semantic segmentation of frames \cite{luc2017}, deep visual image representations of frames \cite{vondrick2016b}, human pose \cite{fragkiadaki2015,butepage2017,martinez2017} and human actions \cite{nguyen2014,kitani2012}.  Others have considered mid-level representation outputs, such as optical flow \cite{yuen2010,walker2015}.  At the lowest and most general level, a growing body of research has explored predicting the pixel intensities of future frames \cite{ranzato2014,srivastava2015,mathieu2016,vondrick2016,villegas2017,vondrick2017,liu2017,kalchbrenner2017,liang2017,walker2017,lotter2017,Cricri2016}. In this paper, we propose a method to predict frame-wise pixel intensities by recurrently transforming image representations into the future.

\textbf{Prediction and transformation} A key differentiating aspect between prior work is the generative process.  Inspired by encoder-decoder language models (e.g. \cite{Sutskever2014}), \cite{ranzato2014,srivastava2015} consider a recurrent network that encodes an input sequence into a fixed length vector and a subsequent recurrent network that decodes the vector to progressively generate predicted frames.  Others have considered a more direct approach that predicts future frames from observed frames using a convolutional neural network (CNN) \cite{mathieu2016,luc2017}.
Several other works have proposed copying or applying simple transformations to past pixels to generate image frame predictions \cite{brabandere2016,finn2016,liu2017,vondrick2017,PatrauceanHC16}. In contrast, we predict future images by transforming and decoding the latent space, rather than by directly predicting future frames or by copying or transforming past pixels.

\textbf{Factored representations} Another line of recent work has approached the problem of prediction by factoring the  representation of the latent information or shaping the latent to learn properties useful for prediction. Vondrick et al.\ \cite{vondrick2016} factor the generative process into separate foreground and background streams that are combined
to create the final video. Goroshin et al.\ \cite{goroshin2015} train a linearized latent space so that future prediction can be treated as linear interpolation. Several works \cite{walker2017,villegas2017b} have considered predicting human pose and then conditioning image generation on the predicted pose. These models can achieve impressive results, but they assume latents with known structure (i.e.\ 2D poses) and are thus limited to human-focused imagery. Our method assumes only that learned representations can be transformed by a learned operator and is not restricted to settings where the latent space can be explicitly labeled.

\textbf{Predicting with motion and content} Most similar to our work are two approaches that factor the latent information to capture scene appearance and dynamics \cite{denton2017,villegas2017}. Denton et al.\ \cite{denton2017} learn separate representations of content and pose and predict future frames by fixing the content and estimating the future pose. This model produces very stable predictions, but it assumes content does not change over a sequence. This limits its applicability to scenes with camera motion and dynamic content. Our method does not assume fixed scenes, but instead learns a representation of state and motion that is designed to accommodate a variety of transformations while still preserving image structure. 

Villegas et al.\ \cite{villegas2017} learn representations of content and motion by feeding two networks with images and difference images, respectively. They train their method with an adversarial loss and need to re-encode predictions to generate more than one future frame. While their method incorporates motion into the representation by splitting the input into images and difference images and directly predicting next frames, our method learns to represent both states and transformations from frames and learns motion by applying transformations to states. Our method produces good results without adversarial training or image re-encoding.

\textbf{Loss functions} Previous work has proposed to improve future prediction by designing loss functions to guide learning to better solutions. While earlier work uses standard pixel-wise reconstruction losses like the mean-squared error (MSE) or binary cross entropy (BCE) loss function \cite{srivastava2015}, more recent work (e.g., \cite{mathieu2016,vondrick2017,lu2017,zeng2017}) often incorporates some form of generative adversarial network (GAN) model \cite{goodfellow2014}, either alone 
or in conjunction with a reconstruction loss, such as MSE. While GANs can produce crisp predictions, they are notoriously hard to train and model convergence is difficult to evaluate \cite{Goodfellow17}. In this paper, we demonstrate that a simple MSE loss is capable of generating good predictions when paired with an appropriately structured architecture.

\section{Technical approach}
In future prediction, we are given a sequence of $T$ images $\{I_1, ..., I_T\}$ and want to produce the most likely sequence of $K$ future images $\{I_{T+1}, ..., I_{T+K} \}$. We seek to do so by capturing how the structure of the image transforms over time. 

Images are high dimensional but the pixel space dimension does not reflect the intrinsic dimensionality of the scene. For example, a $64 \times 64$ image of two translating digits lives in a pixel space of the same dimensionality as a $64 \times 64$ image of a walking person. However, the latter image contains scene content with many more degrees of freedom so its intrinsic dimensionality is higher. Similarly, a $128 \times 128$ and a $64 \times 64$ image of the same walking person both depict content with the same degrees of freedom (up to appearance details lost in downsampling), but with very different pixel dimensions. When we predict images, we must predict pixels, but we seek to do so by modeling the transformation of the images' content. 

\begin{figure}[t]
\begin{center}
	\epsfig{file=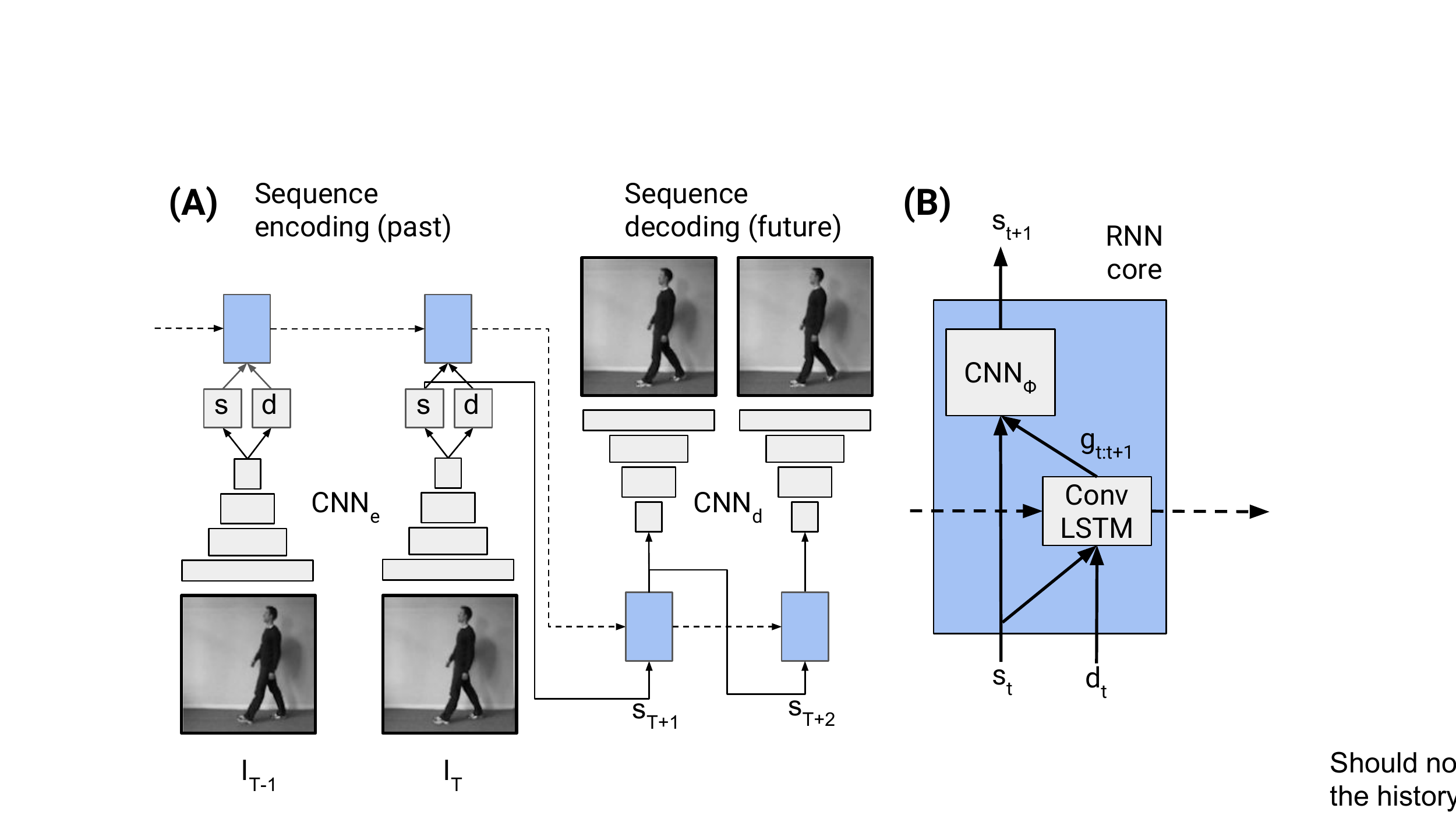,width=4.5in}
\end{center}
\caption{Architecture overview. (A) Our model uses an encoder-decoder sequence-to-sequence architecture with a factorized latent that captures the image state, $s$, and transformation, $d$. Residual connections are omitted for clarity; see the text and Figure \ref{fig:arch_skips} for details. (B) Future states are transformed from past states using an RNN core that accumulates the transformation estimate $g$ with a ConvLSTM and applies it to the recursively estimated state $s$ with an operator $\text{CNN}_{\Phi}$.}
	\label{fig:main_arch}
\end{figure}

Accordingly, we model the instantaneous state of the scene at time $t$ using a latent variable ${s_t}$. Because we do not know the state of future frames, we seek to transform past latent variables $\{s_1, ..., s_T\}$ to estimate the future latents $\{s_{T+1}, ..., s_{T+K}\}$. Future latents depend on previously estimated future latents, so we model this estimation process with a function $f$, such that the estimate at time $k$, where $1 \leq k \leq K$, is given by

\begin{equation}
\hat{s}_t=f(\{ s_1, ..., s_T, \hat{s}_{T+1}, ..., \hat{s}_{T+k-1} \}).
\end{equation}
In the context of image prediction, such a function is typically parameterized with a recurrent neural network (RNN) applied to the output of the encoder of an encoder-decoder architecture \cite{srivastava2015}: 

\begin{equation}
\hat{I}_{T+k} = \text{CNN}_d(\hat{s}_{T+k})
\end{equation}
\begin{equation}
\hat{s}_{T+k} = \text{RNN}(\{ s_1, ..., s_T, \hat{s}_{T+1}, ..., \hat{s}_{T+k-1} \})
\end{equation}
\begin{equation}
s_t = \text{CNN}_e(I_t),
\end{equation}
where $\hat{s}_t$ is the estimated latent state at time $t$, $\hat{I}_t$ is the estimated image at time $t$, and $\text{CNN}_e$ and $\text{CNN}_d$ are encoder and decoder CNNs, respectively. This is illustrated in Figure \ref{fig:main_arch}(A).

While such structures can in principle learn to model arbitrary transformations \cite{siegelmann1995}, these models often struggle to produce realistic image transformations. In practice, these models may learn to memorize transformations as arbitrary mappings from state to state (as illustrated schematically in Figure \ref{fig:overview}(A)) rather than representing transformations as predictable, generalizable mappings like those that characterize the natural transformations between world states.

We now describe how we encourage the network's latent state to learn more predictable mappings by jointly learning representations of state and transformations.

\subsection{Transformational states}

To encourage an encoder-decoder structure to learn to model predictable latent space transformations, we introduce an additional latent variable $d_t$ to capture the evidence available for estimating the transformation from each input image. The output of the encoder CNN at each time step can then be written as 

\begin{equation}{\label{eq:factorization}}
s_t, d_t = \text{CNN}_e(I_t).
\end{equation}
Next, we describe how we encourage the network to exploit the factorization in (\ref{eq:factorization}) by wiring the network so that transformational states $d_t$ cannot directly predict images but must act by transforming states $s_t$. 

We want the $d_t$ to capture all information that is available from $I_t$ about the transformation from state $s_t$ to $s_{t+1}$. Let us call this transformation $g_{t:t+1}$. While individual images provide some information about how the world will move, in general they are insufficient to model the full transformation from $t$ to $t+1$ and to later points in the future. 

To see this, consider the person in image (i) of Figure \ref{fig:overview}(A). From this world state, transformations in time are unlikely to produce image (v), which shows the same person in a different scene, or image (vi), which shows a different person in a similar position. However, the person in image (i) might move his arms closer together (producing image (ii)) or further apart (producing image (iv)). Given image (i) and (ii), however, image (iii) becomes much more likely than image (iv).

That is, the transformation that can be estimated from a single image ($d_t$) will not in general be equivalent to the true state transformation ($g_{t:t+1}$). However, some information about the transformational state of the world is observable from a single image, and we can arrive at better estimates by integrating transformation estimates over time. Accordingly, we use an RNN to incorporate the history of instantaneous transformation estimates $d_t$ and the accompanying states $s_t$ to obtain a better estimate of the transformation $g_{t:t+1}$ acting on $s_t$:
\begin{equation}
g_{t:t+1} = \text{RNN}(\{[d_1, s_1], ..., [d_t, s_t]\}).
\end{equation}
We then model the action of this transformation on latent states as
\begin{equation}
s_{t+1} = \Phi(g_{t:t+1}, s_t),
\end{equation}
where $\Phi$ is an operator that transforms $s_t$ by applying $g_{t:t+1}.$ We parameterize $\Phi$ with a three-layer CNN (with no recurrence), and we parameterize the RNN with a three-layer convolutional long short-term memory (convLSTM) model. We show the full recurrent core, including the CNN operator and  transformation RNN in Figure \ref{fig:main_arch}(B).


We next describe how we integrate skip connections into the model to encourage long-term stability and fidelity of image production while the state is undergoing transformations.

\begin{figure}[t]
\begin{center}
	\epsfig{file=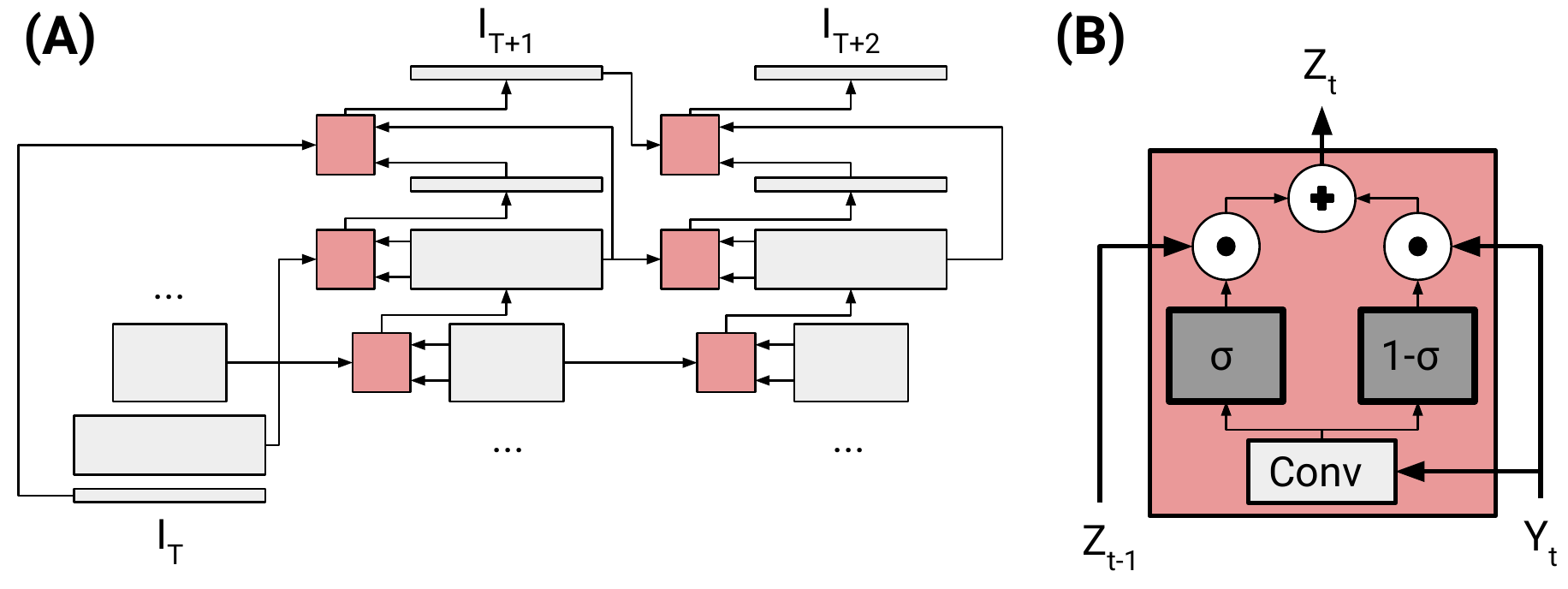,width=4.5in}
\end{center}
\caption{Weighted residual connections. (A) To produce high quality images at multiple time steps in the future without re-encoding images, we use a residual connection scheme designed to gradually alter image content from the last observed input image. Residual connections connect the encoder at time $T$ (last input) to the decoder at time $T+1$ (first output). At subsequent times, the decoder inherits information about the past from the decoder at the previous time only. The network has this connectivity pattern at every layer: we show only two decoder layers and the output image for easier visualization. (B) We use a retinotopic weighting scheme to allow each layer of a decoding network to selectively incorporate skipped input from the past. Weights and feature maps at time $t$ are functions of the predicted latent state $\hat{s}_t$ at time $t$.}
	\label{fig:arch_skips}
\end{figure}

\subsection{Weighted residual connections}

\begin{figure}[!h]
\centering
   \includegraphics[scale=1]{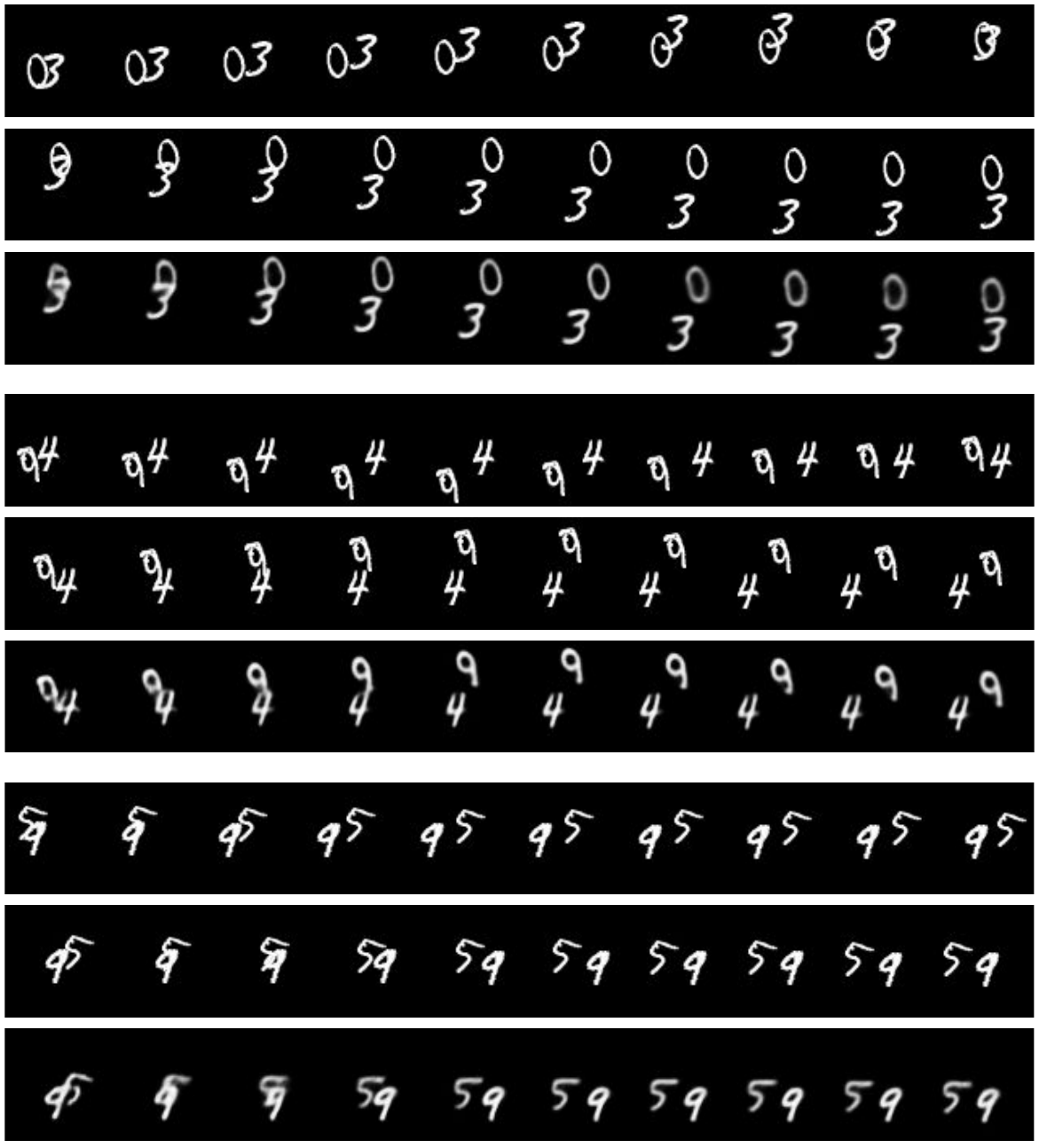}
   \caption{Example sequences on Moving MNIST. For all three examples, the first row shows the input sequence (past), the second row shows the ground truth future, and the third row shows the predicted sequence. Our model is able to stably predict digits over multiple timesteps, even when digits overlap for multiple frames.}
   \label{fig:mnist}
\end{figure}

Recent works \cite{denton2017,villegas2017,finn2016} have found that skip connections from encoder to decoder networks are essential for producing high quality image outputs, especially for capturing high-frequency information of textures and background.  However, when encoded images are in the past and decoded images are in the future, this paradigm is limited in several ways. First, future encoder states cannot be used as a source of skipped image information without first re-encoding estimated images. This may lead to difficulties in CNN/RNN training because of mismatched statistics between true and estimated frames. Second, skipping from past states to future ones can introduce artifacts when static features are copied as if nothing in the scene has changed. This can result in ghosting artifacts that are difficult for the network to learn to correct.

We introduce a mechanism for passing information forward from the encoder state at the last input time step to the decoder at future time steps without re-encoding predictions and without repeatedly copying activations from the past (Figure \ref{fig:arch_skips}). Instead of copying activations from the encoder to the decoder at all future times, we connect the encoder at the last input time step only to the decoder at the first prediction time step. Subsequent decoder time steps take the activations of the decoder at the previous time step and re-weight them. This configuration allows features to flow forward in time from the last input time step, while allowing features to change as necessary to reflect motion and without requiring images to be re-encoded.

The initial feature map $Y^l_t$ output by layer $l$ of the decoder network at time $t$ is combined with skipped output $Z^l_{t-1}$ from the previous time step in the form of a weighted residual connection: 

\begin{equation}
Z^l_t = (1-\sigma(W^l)) \odot Y^l_t + \sigma(W) \odot Z^l_{t-1},
\end{equation}
where $\odot$ denotes element-wise multiplication. $Z^l_t$ is the final output of the network at layer $l$ at time $t$. $W^l_t$ (a weight map) is the output of a $1 \times 1$ convolution with $Y^l_t$ as input. We output one weight value for each spatial position of the feature map and broadcast the weight to all channels to perform the elementwise multiplication. This weighting strategy introduces only $K^l +1$ parameters per layer, where $K^l$ is the number of channels in $Y^l$. 

For the first prediction time step, the skip inputs $Z^l_{t-1}$ come from the layer of the encoder network at the last input time step with matching spatial dimension. Otherwise, they come from the corresponding layer in the decoder at time $t-1$. The weighting scheme is shown in Figure \ref{fig:arch_skips}(A). This configuration is similar to the one introduced in \cite{DBLP:journals/corr/SrivastavaGS15}, applied at each time step. 

\begin{table}[t]
\begin{center}
\begin{tabular}{|c|c|c|}\hline
Model & average, 10 predicted frames & first frame prediction\\\hline
ConvLSTM \cite{shi2015} & 367.2 & - \\
Encoder-Decoder LSTM \cite{srivastava2015} & 341.2 & - \\
Dynamic Filter Networks \cite{brabandere2016} & 285.2 & - \\
Spatiotemporal Autoencoder \cite{PatrauceanHC16} & - & 179.8 \\
Video Pixel Networks \cite{kalchbrenner2017} & 87.6 & - \\
Video Ladder Networks \cite{Cricri2016} & 187.7 & - \\
Ours & 210.1 & 172.4 \\\hline
\end{tabular}
\end{center}
\caption{Comparison of binary cross-entropy (BCE) results (nats/frame) on the Moving MNIST test set. Lower scores indicate better performance.}
\label{table:MNIST}
\end{table}

When paired with our network architecture, this skip configuration allows us to estimate future images without re-encoding estimated images into the encoder CNN. Because subsequent time steps inherit the activations of the previous decoder state, and do not directly copy the states of the last encoder (as in e.g. \cite{2018arXiv180207687D}), we observed that these networks trained more quickly and resulted in fewer ghosting artifacts. 

We incorporate a similar weighted residual scheme to directly skip the last input image to future timesteps. As with feature maps, for all times $t > T+1$ we skip the image from the previous timestep $t-1$ instead of the last input image $I_T$. Directly skipping the final input image to later time steps resulted in lower quality outputs (see Supplemental Figures \ref{fig:kth_ablation_0} and \ref{fig:kth_ablation_1} for examples). We also observed that the residual connection works best when the weighting is applied after the tanh nonlinearity in both images. Weighting before the output nonlinearity led to saturated images at later prediction time steps.

\section{Experiments}
\subsection{Datasets}

\begin{table}[t]
\begin{center}
\begin{tabular}{|c|c|c|c|c|}\hline
\multirow{2}{*}{Model} &\multicolumn{2}{c|}{PSNR} & \multicolumn{2}{c|}{SSIM}\\\cline{2-5}
 & at time 1& average over 10 frames &at time 1& average over 10 frames \\\hline
ConvLSTM \cite{villegas2017} & 33.8&27.6& 0.95& 0.84\\
MCNet \cite{villegas2017} & 33.8&28.2& 0.95& 0.86\\
Ours &34.8&29 & 0.95&0.86\\\hline
\end{tabular}
\end{center}
\caption{Comparison of frame prediction results on the KTH test set. Higher scores indicate better performance.}
\label{table:KTH}
\end{table}

We performed experiments on three datasets: a standard synthetic dataset, Moving MNIST \cite{srivastava2015}, and two real world datasets, KTH actions \cite{KTH} and UCF101 \cite{UCF101}. Moving MNIST is a dataset of synthetic videos, with an arbitrarily large training set (videos are generated procedurally) and a test set of 10,000 videos.  Each video has an image resolution of $64\times 64$ and is 20 frames in length.  The videos capture two digits moving in random directions and with random velocities.  KTH consists of 2391 videos capturing six human actions: boxing,
hand clapping, hand waving, jogging, running, and walking.
As is standard practice
in prior work on frame prediction using KTH, we convert the images to $128 \times 128$ 
prior to processing. All sequences contain scenes with relatively homogeneous backgrounds.  The scenes were captured with a static camera at 25 frames per second.
UCF101 contains 13,320 YouTube videos capturing 101 human actions.  
As done in previous evaluations using UCF101, we convert the images to $256 \times 256$ prior to processing.  Notably, many UCF101 videos contain spatial and temporal (i.e.\ duplicate frames) artifacts due to compression.

\begin{table}[t]
\begin{center}
\begin{tabular}{|c|c|c|}\hline
Model & PSNR & SSIM\\\hline
EpicFlow \cite{revaud2015} & 29.1 & 0.91\\
NextFlow \cite{sedaghat16} & 29.9 & - \\
BeyondMSE \cite{mathieu2016} & 28.2 & 0.89\\
DVF \cite{liu2017} & 29.6 & 0.92\\
Dual Motion GAN \cite{liang2017}& 30.5 & 0.94\\
Ours & 28.3 & 0.88\\\hline
\end{tabular}
\end{center}
\caption{Comparison of next frame prediction results on the UCF101 test set (split 1). Higher scores indicate better performance.}
\label{table:UCF}
\end{table}

\subsection{Architecture and training details}

\begin{figure}[!h]
\centering
\vspace{-0cm}
    
    
    \subfloat{\includegraphics[width=\fsSm\textwidth]{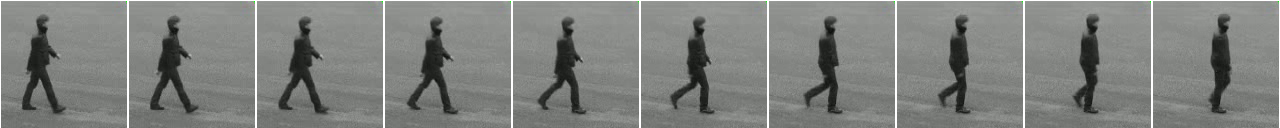}}
    
    \subfloat{\includegraphics[width=\fsBg\textwidth]{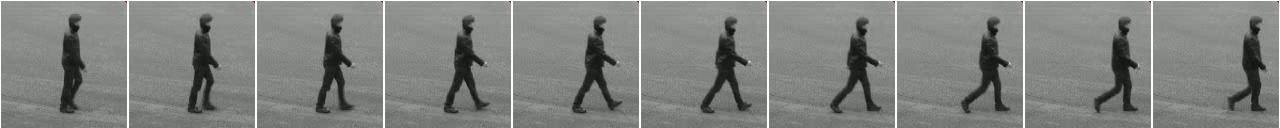}}

	\subfloat{\includegraphics[width=\fsBg\textwidth]{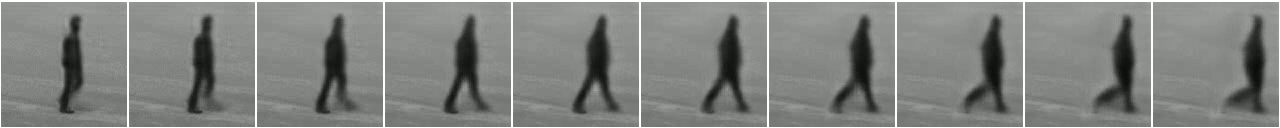}} 
    
    {\centering (A) Walking}
    
\subfloat{{\includegraphics[width=\fsSm\textwidth]{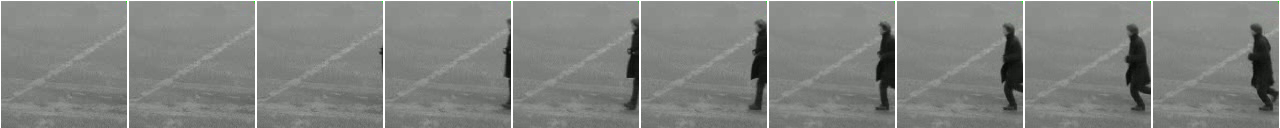}}}
    
\subfloat{{\includegraphics[width=\fsBg\textwidth]{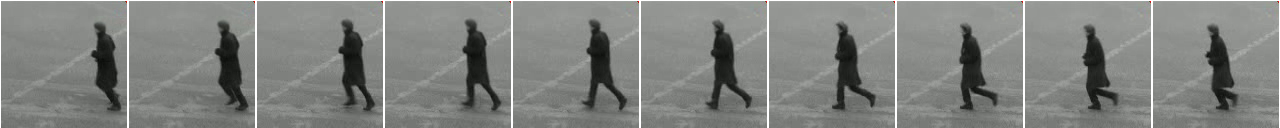}}}
    
    \subfloat{{\includegraphics[width=\fsBg\textwidth]{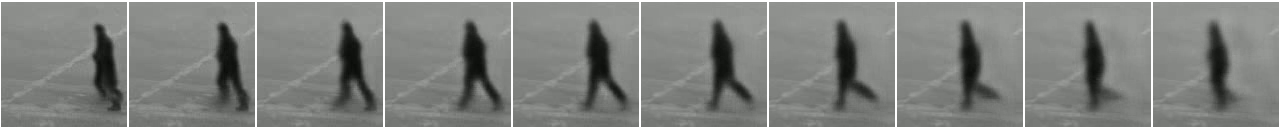}}}
    
    {\centering (B) Running}
    
    \subfloat{{\includegraphics[width=\fsSm\textwidth]{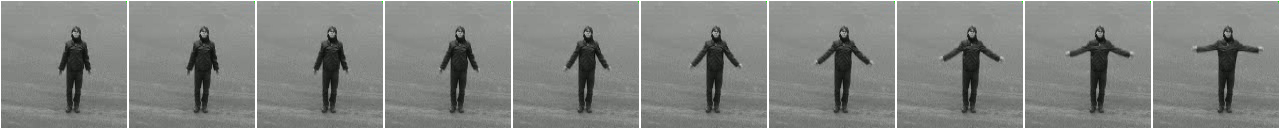}}}
    
    \subfloat{{\includegraphics[width=\fsBg\textwidth]{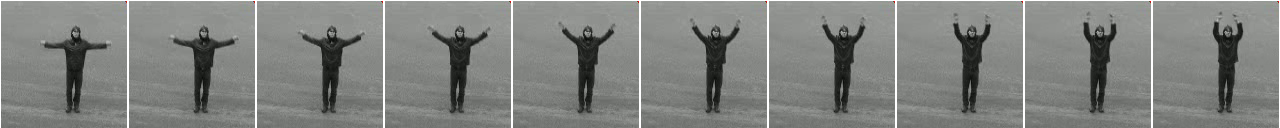}}}
    
    \subfloat{{\includegraphics[width=\fsBg\textwidth]{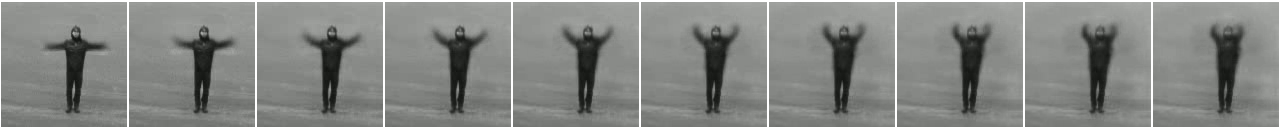}}}
    
    {\centering (C) Hand waving}
    
   \caption{Example sequences on KTH. For all three examples, the first row shows the input sequence (past), the second row shows the ground truth future, and the third row shows the predicted sequence. The model produces faithful motion in a variety of settings and is able to paint in the background after dis-occlusion.}
   \label{fig:kthgood}
\end{figure}

\begin{figure}[t]
\centering
    
    
    

    

        \subfloat{{\includegraphics[width=\fsBgUCFst\textwidth]{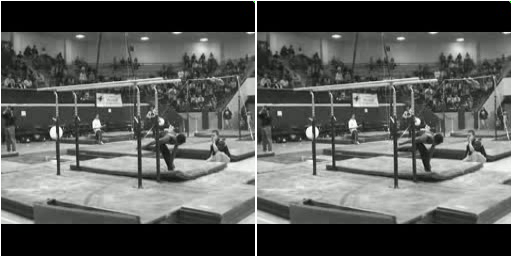}}}~
    \subfloat{{\includegraphics[width=\fsSmUCFst\textwidth]{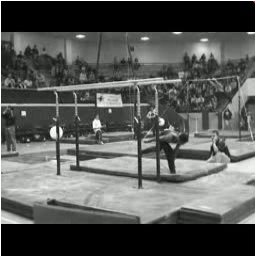}}}~~
    \subfloat{{\includegraphics[width=\fsSmUCFst\textwidth]{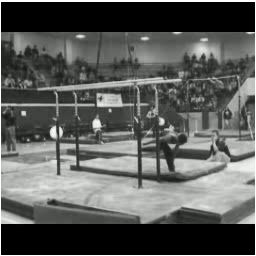}}}~~

    \subfloat{{\includegraphics[width=\fsBgUCFst\textwidth]{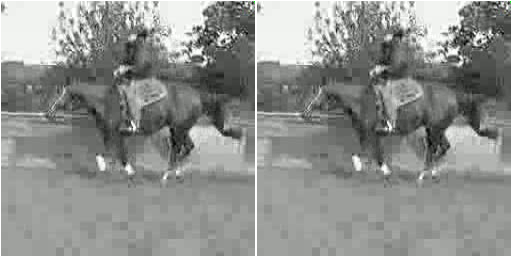}}}~
    \subfloat{{\includegraphics[width=\fsSmUCFst\textwidth]{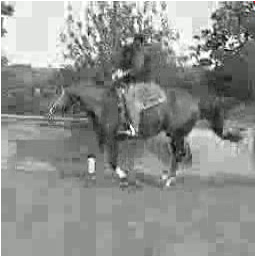}}}~~
    \subfloat{{\includegraphics[width=\fsSmUCFst\textwidth]{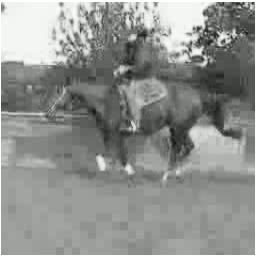}}}~~
    
   \caption{Example sequences on UCF101. For each example, we show two frames from the past followed by the ground truth third frame and the third frame predicted by the model from the first two images.}
   \label{fig:ucf}
\end{figure}

The Moving MNIST and KTH networks were trained to predict 10 frames given 10 input frames and UCF101 networks were trained to predict 1 frame given 2 input frames (to allow us to compare to the compendium of state-of-the-art results in \cite{liang2017}). 
On all datasets probed, we trained the network end-to-end using an average pixel-wise reconstruction loss between the estimated sequences and ground truth future sequences. We use an MSE loss for KTH and UCF1010 and a BCE loss for Moving MNIST. All networks were trained using stochastic gradient descent (SGD) with momentum. We used a starting learning rate of 1 on KTH and UCF101 and 10 on Moving MNIST. We decayed learning rates by a factor of 10 every time the validation loss reached a plateau, until convergence. We used a momentum value of $\beta=$ 0.5 in all cases. We used a weight decay of \num{1e-4} for encoder and decoder weights on all networks, and we included dropout with a rate of 0.5 in all hidden layers of encoder networks on UCF101 and KTH. 

We used horizontal mirroring and random cropping for data augmentation on both UCF101 and KTH datasets. We trained on Moving MNIST with 50 sequences per batch, on KTH with 20 sequences per batch, and on UCF101 with 10 sequences per batch.

The convolutional architectures used on all three datasets are based on the DCGAN architecture \cite{Radford2015}. Each layer of the decoder except for the input layer contains the same number of channels as the corresponding layer of the encoder architecture. Because the decoder does not take the transformational latent as input, the decoder input is of size $4 \times 4 \times N_s$, while the encoder output is of size $4 \times 4 \times (N_s + N_s)$, where $N_s$ is the number of channels in the state latent $s$ and $N_d$ is the number of channels in the transformational latent $d$. In all architectures used here, $N_s = N_d$. We did not perform hyperparameter search for the values of $N_s$ and $N_d$ or the architectures used for encoders and decoder CNNs, and it is likely that better results can be obtained using optimized settings. 

The architectures we use on Moving MNIST, KTH, and UCF101 differ only in the number of layers and the number of filters per layer in the encoder and decoder CNNs. Architecture depths were chosen so that the spatial size the encoder output (and decoder input) was $4 \times 4$. We specify full architectures in the supplemental material (supplemental section \ref{section:supp_archs}). We will make the model code and trained models available upon paper acceptance.

\subsection{Evaluation}

It is difficult to quantitatively evaluate prediction results because reconstruction errors and other measures do not  generally fully capture the perceptual quality of reconstructed images \cite{Theis2016a,Wang2004,mathieu2016}. Nonetheless, quantitative evaluations can give a reasonable indication of the average quality of a method when seen alongside the qualitative results the method produces. 

We evaluate our methods using the error measures most commonly used in the literature: binary cross entropy for Moving MNIST \cite{srivastava2015} and peak signal to noise ratio (PSNR) and Structural Similarity (SSIM) \cite{Wang2004} for KTH and UCF101. We evaluate SSIM using a window of 7x7 pixels with uniform weighting (the same parameters as \cite{denton2017}).

We show quantitative results for Moving MNIST in Table \ref{table:MNIST}, KTH in Table \ref{table:KTH}, and UCF in Table \ref{table:UCF}. In all cases, our results are competitive with state of the art. Because of the large number of architectures and loss configurations in the literature, it is infeasible to thoroughly test all architecture and loss configurations. We report results based on the numbers used in the literature. To aid interpretation of our results in the context of the sequence prediction literature, we include a table comparing the different architectural and loss configurations in the supplement (Supplementary Table \ref{arch-comparisons}). 

We show sample qualitative results on the three datasets in Figure \ref{fig:mnist} (Moving MNIST), Figure \ref{fig:kthgood} (KTH), and Figure \ref{fig:ucf} (UCF). Our method produces reasonable results with good motion in many of settings in these three datasets. The output of dynamic models are difficult to evaluate based on static images alone, and consequently the results of our method are best understood by examining the videos on the project website (\url{https://daniilidis-group.github.io/transformational_states}). To aid interpretion of our results, we also show failure cases on KTH in Figure \ref{fig:kthbad}. Additionally, we show prediction results produced by models with network ablations in the supplement: ablations on Moving MNIST are shown in Supplementary Figure \ref{fig:mnist_ablation} and ablations on KTH are shown in Supplementary Figures \ref{fig:kth_ablation_0} and \ref{fig:kth_ablation_1}. Ablation results are shown on random sequences from the test data in all cases for fair comparison.


\begin{figure}[t]
\centering
    \subfloat{{\includegraphics[width=\fsSm\textwidth]{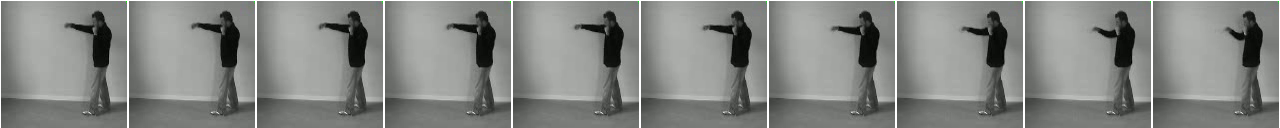}}}~~
    
    \subfloat{{\includegraphics[width=\fsBg\textwidth]{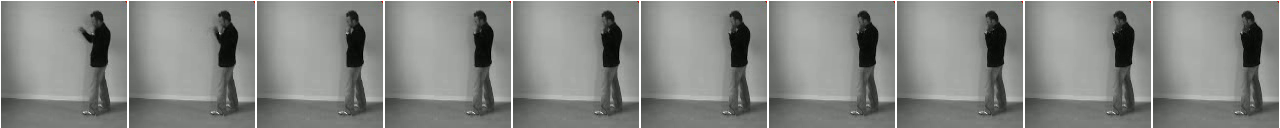}}}~~
    
    
    \subfloat{{\includegraphics[width=\fsBg\textwidth]{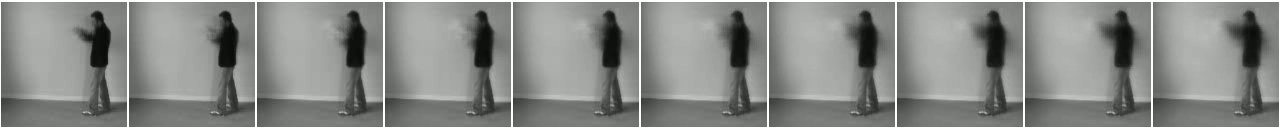}}}~~
    
    {\centering (A) Boxing}
    
    \subfloat{{\includegraphics[width=\fsSm\textwidth]{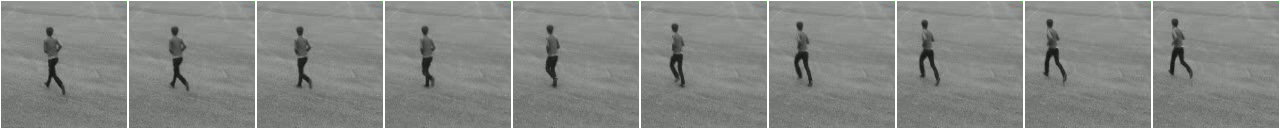}}}~~
    
    \subfloat{{\includegraphics[width=\fsBg\textwidth]{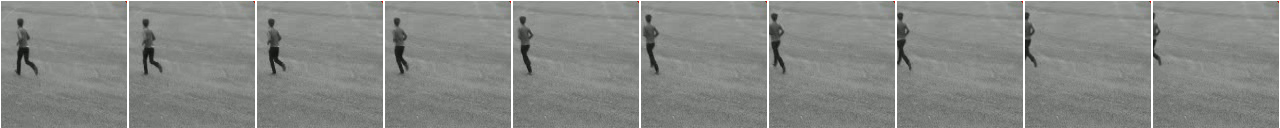}}}~~
    
    \subfloat{{\includegraphics[width=\fsBg\textwidth]
    {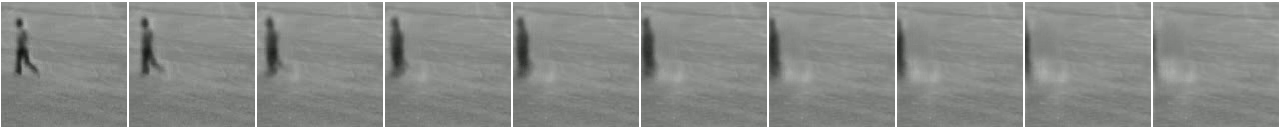}}}~~
    
    {\centering (B) Running}
    
    \subfloat{{\includegraphics[width=\fsSm\textwidth]{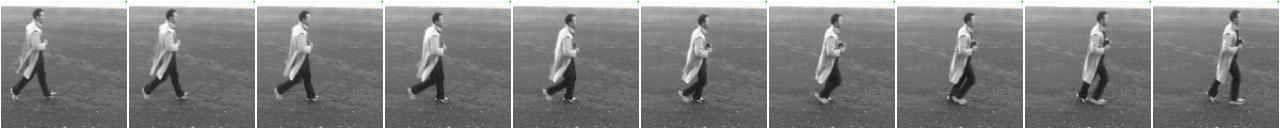}}}~~
    
    \subfloat{{\includegraphics[width=\fsBg\textwidth]{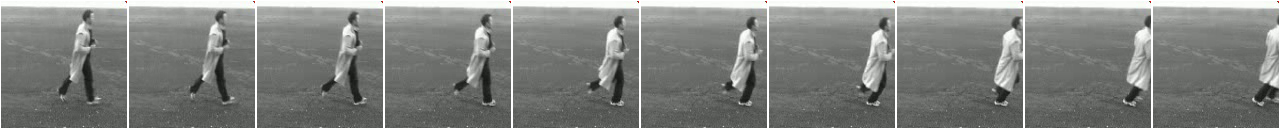}}}~~
    
    
  \subfloat{{\includegraphics[width=\fsBg\textwidth]{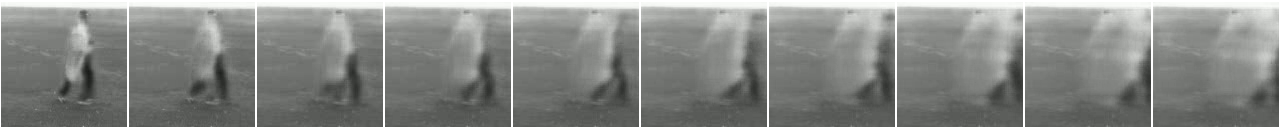}}}~~
  
    {\centering (C) Walking}
  
   \caption{Example failure cases on KTH. (A) The model outputs a blurry motion sequence that does not correspond to the ground truth. (B) The model fails to correctly predict motion or paint in the background when the moving object occupies only a small part of the image. (C) The model fails to correctly paint in the background after the foreground moves, leading to ghosting artifacts.}
   \label{fig:kthbad}
\end{figure}

\section{Summary}

We have described a model for predicting sequences of future images using an architecture that learns latent states and their transformations to future states. We show how to couple this architecture with weighted residual connections from past to future time steps to produce images that are stable after recursive transformations. The resulting network can be trained to predict reasonable results on synthetic and real datasets without requiring direct pixel copying or a GAN. Our model produces good qualitative results and achieves quantitative results comparable to state-of-the-art on several image prediction datasets.

\vskip 0.5cm
\noindent \textbf{Acknowledgments}
\noindent We thank Kenneth Chaney and Nikos Kolotouros for computing support, Stephen Phillips for helpful comments, and the members of the Daniilidis group and the vision community at Penn for many fruitful discussions. We are grateful for support through the following grants: NSF-DGE-0966142 (IGERT),  ARL RCTA W911NF-10-2-0016, and ONR N00014-17-1-2093. K.G.D. is supported by a Canadian NSERC Discovery grant.

\clearpage

\bibliographystyle{splncs}
\bibliography{bibref_definitions_short,bibtex}

\clearpage

\title{Supplemental material: Predicting the Future with Transformational States} 




\authorrunning{Jaegle, Rybkin, Derpanis, and Daniilidis}
\titlerunning{Predicting the Future with Transformational States}
\author{}
\institute{}

\maketitle

\section{Video results}

In videos included on the project website (\url{https://daniilidis-group.github.io/transformational_states}), we visualize prediction results from our model on a large number of sequences from Moving MNIST, KTH, and UCF101. Videos are chosen randomly from the three datasets. In all cases, we show the full sequence given as input to the network (10 frames for Moving MNIST and KTH, 2 frames for UCF101) and the ground truth future sequence along with the network prediction (10 frames for Moving MNIST and KTH, 1 frame for UCF101). Videos are looped and the frames predicted by the network are highlighted in green for clarity.

\section{Network architectures} \label{section:supp_archs}

Full architectures for CNN encoders, CNN decoders, and the two components of the full RNN core ($\text{CNN}_{\Phi}$ and RNN) are given for the three datasets below. Parameters for convolutional (Conv), transposed convolutional (TransposedConv), and convolutional LSTM (ConvLSTM) layers are specified as \{input feature map spatial dimensions, filter size, number of filters, convolution stride\}. Other network elements either have no parameters (tanh and sigmoid activation functions) or always use the default Tensorflow parameter settings (batch norm (BN), leaky rectified linear unit (LReLU)). 

Network components are wired together as in Figure \ref{fig:main_arch}. Weighted residual connections are used identically in the KTH and UCF architectures. The Moving MNIST architecture omits the residual connection to the output image, but is otherwise identical. Note that CNN encoders and $\text{CNN}_{\Phi}$ components include an output tanh nonlinearity to make it easier for the network to match their output distributions to that of a ConvLSTM.

\pagebreak

\noindent \textbf{Moving MNIST:}

\textbf{Encoder}

$\text{Conv}\{64 \times 64, 4 \times 4, 64, 2\} \rightarrow \text{LReLU} \rightarrow$

$\text{Conv}\{32 \times 32, 4 \times 4, 64, 2\} \rightarrow \text{BN} \rightarrow \text{LReLU} \rightarrow$

$\text{Conv}\{16 \times 16, 4 \times 4, 96, 2\} \rightarrow \text{BN} \rightarrow \text{LReLU} \rightarrow$

$\text{Conv}\{8 \times 8, 4 \times 4, 96, 2\} \rightarrow \text{BN} \rightarrow \text{LReLU} \rightarrow$

$\text{Conv}\{4 \times 4, 4 \times 4, 128, 1\} \rightarrow \text{BN} \rightarrow \text{tanh}$

\textbf{Decoder}

$\text{TransposedConv}\{4 \times 4, 4 \times 4, 96, 2\} \rightarrow \text{BN} \rightarrow \text{LReLU} \rightarrow$

$\text{TransposedConv}\{8 \times 8, 4 \times 4, 96, 2\} \rightarrow \text{BN} \rightarrow \text{LReLU} \rightarrow$

$\text{TransposedConv}\{16 \times 16, 4 \times 4, 64, 2\} \rightarrow \text{BN} \rightarrow \text{LReLU} \rightarrow$

$\text{TransposedConv}\{32 \times 32, 4 \times 4, 64, 2\} \rightarrow \text{BN} \rightarrow \text{LReLU} \rightarrow$

$\text{TransposedConv}\{64 \times 64, 4 \times 4, 1, 1\} \rightarrow \text{sigmoid}$

\textbf{RNN}

$\text{ConvLSTM}\{4 \times 4, 3 \times 3, 64, 1\} \rightarrow$

$\text{ConvLSTM}\{4 \times 4, 3 \times 3, 64, 1\} \rightarrow$

$\text{ConvLSTM}\{4 \times 4, 3 \times 3, 64, 1\}$

\textbf{$\text{CNN}_{\Phi}$}

$\text{Conv}\{4 \times 4, 4 \times 4, 64, 1\} \rightarrow \text{LReLU} \rightarrow$

$\text{Conv}\{4 \times 4, 4 \times 4, 64, 1\} \rightarrow \text{LReLU} \rightarrow$

$\text{Conv}\{4 \times 4, 4 \times 4, 64, 1\} \rightarrow \text{tanh}$

\noindent \textbf{KTH:}

\textbf{Encoder}

$\text{Conv}\{128 \times 128, 4 \times 4, 64, 2\} \rightarrow \text{LReLU} \rightarrow$

$\text{Conv}\{64 \times 64, 4 \times 4, 128, 2\} \rightarrow \text{BN} \rightarrow \text{LReLU} \rightarrow$

$\text{Conv}\{32 \times 32, 4 \times 4, 256, 2\} \rightarrow \text{BN} \rightarrow \text{LReLU} \rightarrow$

$\text{Conv}\{16 \times 16, 4 \times 4, 512, 2\} \rightarrow \text{BN} \rightarrow \text{LReLU} \rightarrow$

$\text{Conv}\{8 \times 8, 4 \times 4, 512, 2\} \rightarrow \text{BN} \rightarrow \text{LReLU} \rightarrow$

$\text{Conv}\{4 \times 4, 4 \times 4, 256, 1\} \rightarrow \text{BN} \rightarrow \text{tanh}$

\textbf{Decoder}

$\text{TransposedConv}\{4 \times 4, 4 \times 4, 512, 2\} \rightarrow \text{BN} \rightarrow \text{LReLU} \rightarrow$

$\text{TransposedConv}\{8 \times 8, 4 \times 4, 512, 2\} \rightarrow \text{BN} \rightarrow \text{LReLU} \rightarrow$

$\text{TransposedConv}\{16 \times 16, 4 \times 4, 256, 2\} \rightarrow \text{BN} \rightarrow \text{LReLU} \rightarrow$

$\text{TransposedConv}\{32 \times 32, 4 \times 4, 128, 2\} \rightarrow \text{BN} \rightarrow \text{LReLU} \rightarrow$

$\text{TransposedConv}\{64 \times 64, 4 \times 4, 64, 2\} \rightarrow \text{BN} \rightarrow \text{LReLU} \rightarrow$

$\text{TransposedConv}\{128 \times 128, 4 \times 4, 1, 1\} \rightarrow \text{tanh}$

\textbf{RNN}

$\text{ConvLSTM}\{4 \times 4, 3 \times 3, 128, 1\} \rightarrow$

$\text{ConvLSTM}\{4 \times 4, 3 \times 3, 128, 1\} \rightarrow$

$\text{ConvLSTM}\{4 \times 4, 3 \times 3, 128, 1\}$

\textbf{$\text{CNN}_{\Phi}$}

$\text{Conv}\{4 \times 4, 4 \times 4, 128, 1\} \rightarrow \text{LReLU} \rightarrow$

$\text{Conv}\{4 \times 4, 4 \times 4, 128, 1\} \rightarrow \text{LReLU} \rightarrow$

$\text{Conv}\{4 \times 4, 4 \times 4, 128, 1\} \rightarrow \text{tanh}$

\pagebreak

\noindent \textbf{UCF:}

\textbf{Encoder}

$\text{Conv}\{256 \times 256, 4 \times 4, 64, 2\} \rightarrow \text{LReLU} \rightarrow$

$\text{Conv}\{128 \times 128, 4 \times 4, 128, 2\} \rightarrow \text{BN} \rightarrow \text{LReLU} \rightarrow$

$\text{Conv}\{64 \times 64, 4 \times 4, 256, 2\} \rightarrow \text{BN} \rightarrow \text{LReLU} \rightarrow$

$\text{Conv}\{32 \times 32, 4 \times 4, 256, 2\} \rightarrow \text{BN} \rightarrow \text{LReLU} \rightarrow$

$\text{Conv}\{16 \times 16, 4 \times 4, 512, 2\} \rightarrow \text{BN} \rightarrow \text{LReLU} \rightarrow$

$\text{Conv}\{8 \times 8, 4 \times 4, 512, 2\} \rightarrow \text{BN} \rightarrow \text{LReLU} \rightarrow$

$\text{Conv}\{4 \times 4, 4 \times 4, 512, 1\} \rightarrow \text{BN} \rightarrow \text{tanh}$

\textbf{Decoder}

$\text{TransposedConv}\{4 \times 4, 4 \times 4, 512, 2\} \rightarrow \text{BN} \rightarrow \text{LReLU} \rightarrow$

$\text{TransposedConv}\{8 \times 8, 4 \times 4, 512, 2\} \rightarrow \text{BN} \rightarrow \text{LReLU} \rightarrow$

$\text{TransposedConv}\{16 \times 16, 4 \times 4, 256, 2\} \rightarrow \text{BN} \rightarrow \text{LReLU} \rightarrow$

$\text{TransposedConv}\{32 \times 32, 4 \times 4, 256, 2\} \rightarrow \text{BN} \rightarrow \text{LReLU} \rightarrow$

$\text{TransposedConv}\{64 \times 64, 4 \times 4, 128, 2\} \rightarrow \text{BN} \rightarrow \text{LReLU} \rightarrow$

$\text{TransposedConv}\{128 \times 128, 4 \times 4, 64, 2\} \rightarrow \text{BN} \rightarrow \text{LReLU} \rightarrow$

$\text{TransposedConv}\{256 \times 256, 4 \times 4, 1, 2\} \rightarrow \text{tanh}$

\textbf{RNN}

$\text{ConvLSTM}\{4 \times 4, 3 \times 3, 256, 1\} \rightarrow$

$\text{ConvLSTM}\{4 \times 4, 3 \times 3, 256, 1\} \rightarrow$

$\text{ConvLSTM}\{4 \times 4, 3 \times 3, 256, 1\}$

\textbf{$\text{CNN}_{\Phi}$}

$\text{Conv}\{4 \times 4, 3 \times 3, 256, 1\} \rightarrow \text{LReLU} \rightarrow$

$\text{Conv}\{4 \times 4, 3 \times 3, 256, 1\} \rightarrow \text{LReLU} \rightarrow$

$\text{Conv}\{4 \times 4, 3 \times 3, 256, 1\} \rightarrow \text{tanh}$

\section{Comparison to other prediction models}

In Supplementary Table \ref{arch-comparisons}, we compare details of the architecture and training configurations used in various recently proposed architectures for future prediction. Most notably, we produce future predictions without re-encoding predicted images as input for the encoder network, without directly copying from the input sequence, and without using GANs at any point in network training. The gradient difference loss (GDL) is defined in \cite{mathieu2016}.

We strongly encourage the reader to investigate the cited papers for more details: this table is intended only as a road map to the very interesting and growing literature on future prediction.

\begin{table}[]
\centering
\caption{Comparison of sequence prediction model components and training configurations.}
\begin{adjustbox}{width=1\textwidth}
\label{arch-comparisons}
\begin{tabular}{|l||l|l|l|l|l|l|}
\hline
                                                                   & \begin{tabular}[c]{@{}l@{}}Uses skip connections \\ or copies past?\end{tabular}                                & \begin{tabular}[c]{@{}l@{}}Re-encodes images to \\ generate predictions \\ after t=T+1?\end{tabular} & Uses LSTMs?                                                                                    & Uses additional labels or training?                                                                                                                & Uses GANs?                                                                                                     & Loss                                                                                             \\ \hline \hline
BeyondMSE \cite{mathieu2016}                                                          & \begin{tabular}[c]{@{}l@{}}Uses multi-scale \\ Laplacian pyramid on \\ full sequence (re-)encoding\end{tabular}  & Yes                                                                                                  & No                                                                                             & No                                                                                                                                                 & \begin{tabular}[c]{@{}l@{}}GAN on \\ predicted images\end{tabular}                                             & MSE, GDL, GAN                                                                                    \\ \hline
MCNet \cite{villegas2017}                                                              & \begin{tabular}[c]{@{}l@{}}Skips from previous \\ frame and difference \\ image re-encoding\end{tabular}        & \begin{tabular}[c]{@{}l@{}}Re-encodes images\\ and re-computes\\ difference images\end{tabular}      & \begin{tabular}[c]{@{}l@{}}ConvLSTM on \\ difference image \\ encoding\end{tabular}            & No                                                                                                                                                 & \begin{tabular}[c]{@{}l@{}}GAN on \\ predicted images\end{tabular}                                             & MSE, GDL, GAN                                                                                    \\ \hline
DRNet \cite{denton2017}                                                              & \begin{tabular}[c]{@{}l@{}}Copies content vector\\ from last time step\end{tabular}                             & No                                                                                                   & \begin{tabular}[c]{@{}l@{}}LSTM on input \\ sequence embedding\end{tabular}                    & \begin{tabular}[c]{@{}l@{}}Encoder output trained to \\ disentangle content from pose, \\ content to remain static over \\ a sequence\end{tabular} & \begin{tabular}[c]{@{}l@{}}GAN to \\ disentangle content \\ and pose\end{tabular}                              & \begin{tabular}[c]{@{}l@{}}Two-stage training: \\ (1) GAN, \\ (2) MSE\end{tabular}               \\ \hline
SVG-LP \cite{2018arXiv180207687D}                                                             & \begin{tabular}[c]{@{}l@{}}Skips from last \\ input frame encoding\end{tabular}                                 & Yes                                                                                                  & \begin{tabular}[c]{@{}l@{}}LSTM on \\ encoder output and \\ LSTM on learned prior\end{tabular} & No                                                                                                                                                 & No                                                                                                             & \begin{tabular}[c]{@{}l@{}}MSE, KL divergence \\ between model and \\ learned prior\end{tabular} \\ \hline
SNA \cite{ebert2017}                                                                & \begin{tabular}[c]{@{}l@{}}Skips from previous \\ frame re-encoding and \\ from last input frame\end{tabular}   & Yes                                                                                                  & \begin{tabular}[c]{@{}l@{}}ConvLSTM layers \\ throughout encoder \\ and decoder\end{tabular}   & \begin{tabular}[c]{@{}l@{}}Uses control state and \\ action as additional input\end{tabular}                                                       & No                                                                                                             & MSE                                                                                              \\ \hline
Dual Motion GAN \cite{liang2017}                                                    & No                                                                                                              & Yes                                                                                                  & \begin{tabular}[c]{@{}l@{}}ConvLSTM on \\ encoder output\end{tabular}                          & \begin{tabular}[c]{@{}l@{}}Trains network to \\ predict current and \\ future optical flow\end{tabular}                                            & \begin{tabular}[c]{@{}l@{}}GAN on \\ predicted images and \\ predicted current and \\ future flow\end{tabular} & \begin{tabular}[c]{@{}l@{}}GAN, VAE KL \\ divergence\end{tabular}                                \\ \hline
DVF \cite{liu2017}                                                                & No                                                                                                              & Yes                                                                                                  & No                                                                                             & No                                                                                                                                                 & No                                                                                                             & L1, total variation losses                                                                       \\ \hline
PredNet \cite{lotter2017}                                                            & No                                                                                                              & Yes                                                                                                  & \begin{tabular}[c]{@{}l@{}}ConvLSTMs throughout\\ architectures\end{tabular}                   & No                                                                                                                                                 & No                                                                                                             & \begin{tabular}[c]{@{}l@{}}Predictive coding \\ L1 loss\end{tabular}                             \\ \hline
\begin{tabular}[c]{@{}l@{}}Adversarial \\ Transformer \cite{vondrick2017} \end{tabular} & \begin{tabular}[c]{@{}l@{}}Predicted images given\\ as interpolation of \\ last input image pixels\end{tabular} & \begin{tabular}[c]{@{}l@{}}No, last input image\\ directly transformed\end{tabular}                  & No                                                                                             & No                                                                                                                                                 & \begin{tabular}[c]{@{}l@{}}Uses conditional GAN \\ on predicted sequences\end{tabular}                         & GAN                                                                                              \\ \hline
Ours                                                               & \begin{tabular}[c]{@{}l@{}}Masked residual from \\ previous decoder state\end{tabular}                          & No                                                                                                   & \begin{tabular}[c]{@{}l@{}}ConvLSTM on \\ encoder output\end{tabular}                          & No                                                                                                                                                 & No                                                                                                             & MSE                                                                                              \\ \hline
\end{tabular}
\end{adjustbox}
\end{table}

\section{Ablation studies}

\begin{figure}[!h]
\centering
   \includegraphics[scale=0.7]{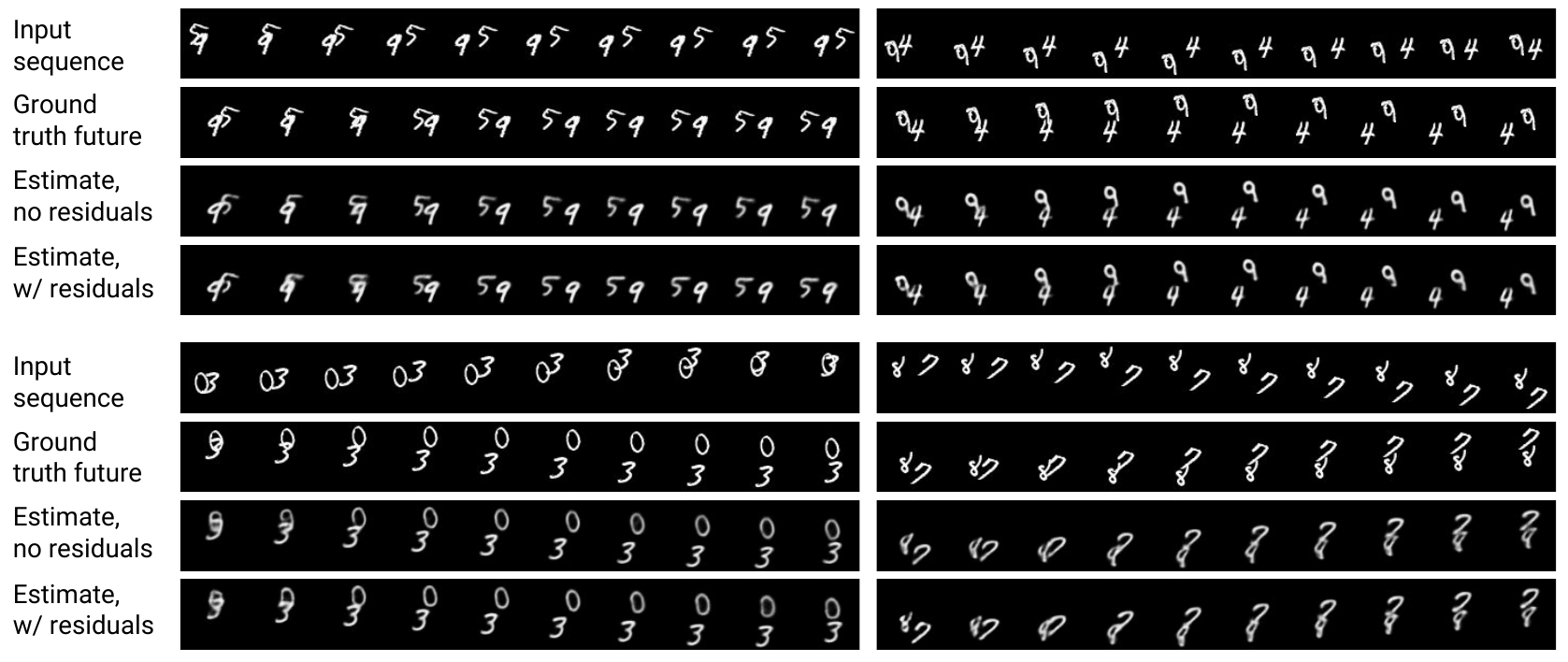}
   \caption{Comparison of Moving MNIST results on architectures with and without residual connections. The model labeled ``no residuals" has no skip or residual connections of any kind. The model labeled ``w/ residuals" is the full model described in the paper. The model without weighted residual connections produces good predictions, but including these connections produces crisper results, especially at early prediction time steps. Both architectures reliably capture digit identity, even after the digits overlap.}
   \label{fig:mnist_ablation}
\end{figure}

\begin{figure}[!h]
\centering
   \includegraphics[scale=0.8]{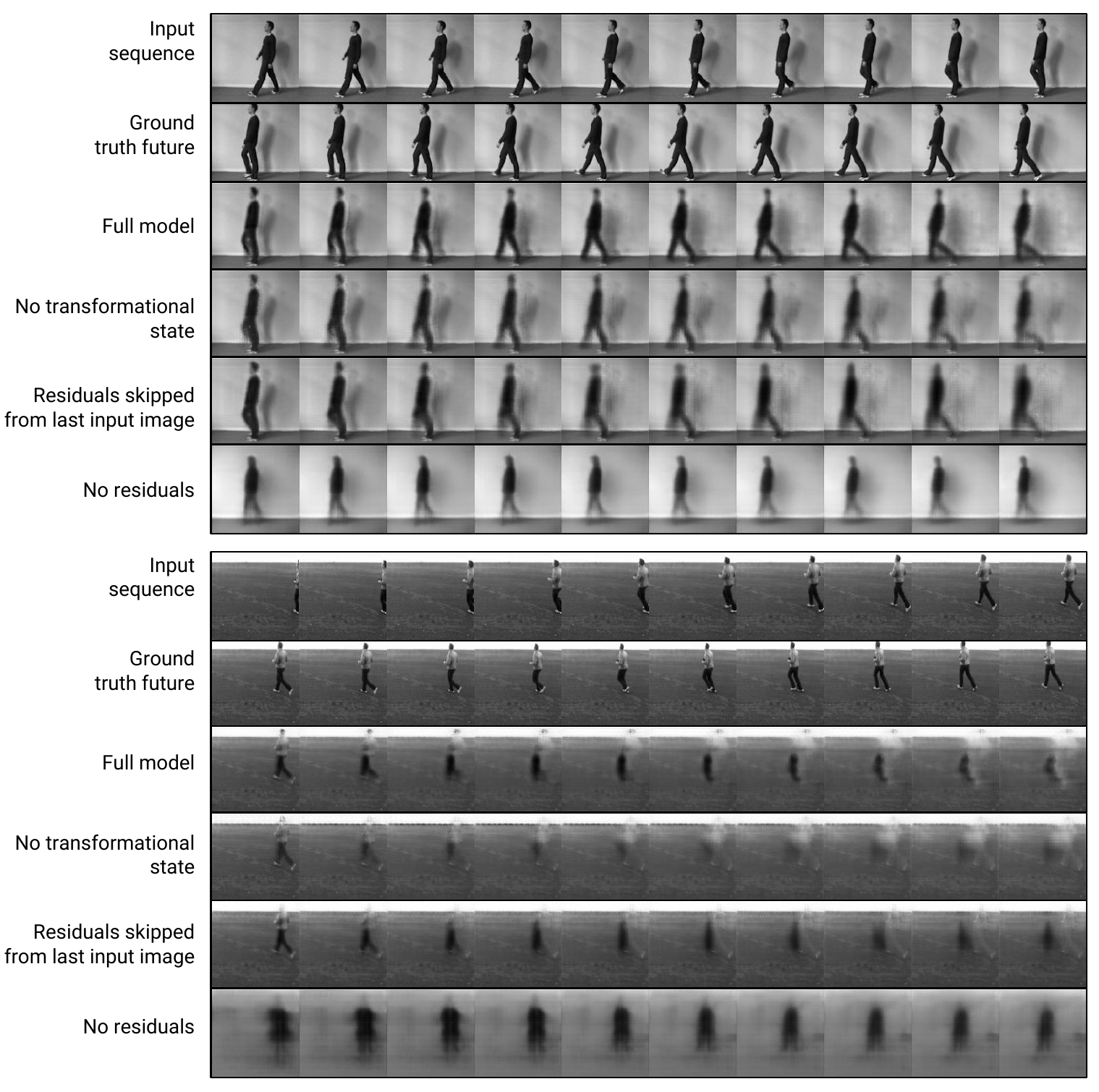}
   \caption{Comparison of KTH results on models with architectural ablations. (i) ``No transformational state'': the RNN core omits the $\text{CNN}_{\Phi}$ and includes only ConvLSTM components. (ii) ``Residuals skipped from last input image'': each decoder directly receives residual input from the encoder at the last input time step ($t=T$) instead of the previous decoder time step. Weighted residuals are still used. (iii) ``No residuals": no residual or skip connections of any kind are used. The second sequence shown here is very challenging for all models. While not perfect, the full model produces better motion (notice the motion of the legs) and less prominent ghosting artifacts than ablations.}
   \label{fig:kth_ablation_0}
\end{figure}

\begin{figure}[!h]
\centering
   \includegraphics[scale=0.8]{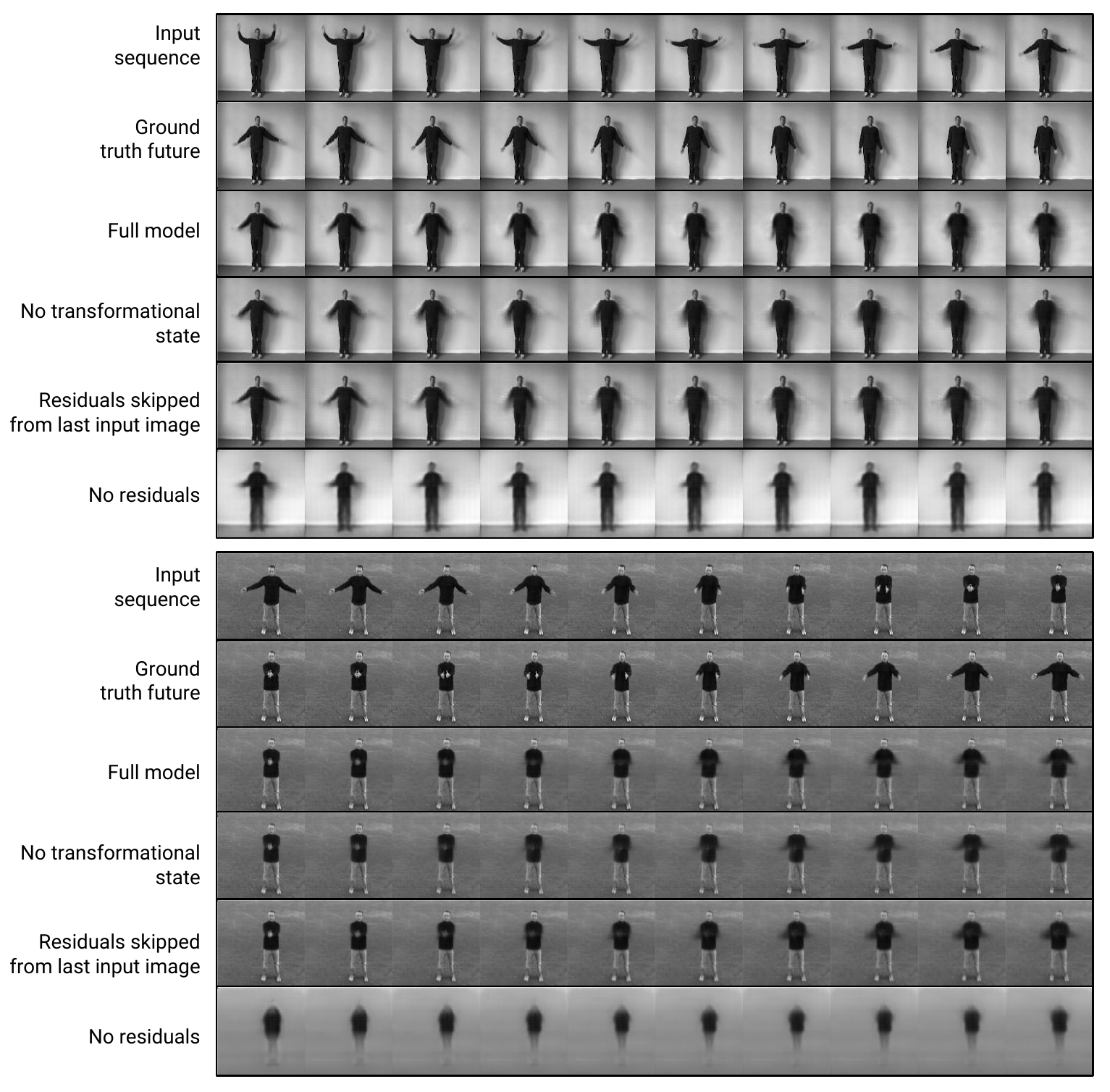}
   \caption{Additional comparisons of KTH results on models with architectural ablations. See Supplemental Figure \ref{fig:kth_ablation_0} caption for explanation of ablations.}
   \label{fig:kth_ablation_1}
\end{figure}

Here, we present qualitative results from ablations of the proposed architecture on Moving MNIST (Supplemental Figure \ref{fig:mnist_ablation}) and KTH (Supplemental Figures \ref{fig:kth_ablation_0} and \ref{fig:kth_ablation_1}). On Moving MNIST, we show the results of training the model with and without weighted residual connections. Moving MNIST images are fairly simple, so residual connections do not lead to as large an improvement in performance as on datasets with real-world image statistics.

On KTH, we compare the full model against models trained (i) with a ConvLSTM core instead of the full RNN core described in the main paper, (ii) using residual connections directly from the last input time step instead of the previous time step (i.e. the decoder at time $t=T+k$ receives skip connections from the encoder at time $t=T$ instead of the decoder at time $t=T+k-1$), (iii) with no skip or residual connections. The full model best captures image motion while also leading to better background in-painting.

All models were trained with the hyperparameters used to train the model described in the paper. On KTH, this produced good results for all models including residual connections. We have seen qualitatively better motion on the model without residuals using different hyperparameters, but the results shown here are representative of the difference between models. Including residual connections led to dramatically better results on background prediction, but the model without residual connections appears to model motion reasonably well in some cases. 

\end{document}